\documentclass[sigconf]{acmart}

\usepackage{booktabs}
\usepackage{amsmath}
\usepackage{xspace}
\usepackage{pgfplots}
\pgfplotsset{compat=1.17}
\usetikzlibrary{decorations.pathreplacing}

\newcommand{\ic}{\mathrm{IC}\xspace}
\newcommand{\forgetic}{\ensuremath{\ic_{\mathcal{F}}}\xspace}
\newcommand{\Lzero}{\textsf{L0}\xspace}
\newcommand{\Lone}{\textsf{L1}\xspace}
\newcommand{\Ltwo}{\textsf{L2}\xspace}
\newcommand{\Lthree}{\textsf{L3}\xspace}
\newcommand{\gapclosed}{\ensuremath{\kappa}\xspace}

\AtBeginDocument{%
  }

\setcopyright{none}
\renewcommand\footnotetextcopyrightpermission[1]{}
\begin{document}

\title[Can You Delete a Year of Market Data?]{Can You Delete a Year of
Market Data?\\ Machine Unlearning Against Exact Retraining Oracles}


\author{Junyi Ye}
\orcid{0000-0002-6348-5207}
\affiliation{%
  \institution{School of Computing, Montclair State University}
  \city{Montclair}
  \state{New Jersey}
  \country{USA}
}
\email{yej@montclair.edu}
\begin{abstract}
When a data license expires, deleting stored records does not remove
influence encoded in a trained forecaster. Machine unlearning seeks to
remove this influence without retraining. We benchmark temporal
unlearning with 3{,}200 paired references trained on all data and oracles
retrained without the requested period. The grid covers five
architectures, four rolling folds, five deletable years, and three
experimental deletion levels on an S\&P~500 volatility panel. The 2020
COVID crisis year produces the largest memorization gap for every
architecture. Removing it improves all three deployable models in every
fold, with the largest improvement in the 2022 bear market, while the
two non-deployable models respond inconsistently. The target for
approximate unlearning is the oracle, not low predictive accuracy on the
deleted period. In one Transformer cell, an oracle that never trained on
2020 still predicts it at an information coefficient of $0.51$, compared
with $0.55$ for the reference; pushing predictions toward noise reduces
test skill. Across twelve deployable architecture--method pairs, only
TSMixer with the hinge method remains near the oracle in every fold,
closing $74$--$118\%$ of the reference-to-oracle gap without a measurable
loss of test skill. Method rankings vary across architectures and
rolling windows. Audit separation rises with prior memorization but can
remain small after exact deletion. The window-level loss comparison
reaches at most $0.69$, and treating stock-level windows as independent
inflates the absolute $t$-statistic by a median factor of $1.9$. These
results call for an explicit deletion scope, oracle validation for the
relevant architecture and window, and power-aware auditing.
\end{abstract}






\pagestyle{plain}

\maketitle

\section{Introduction}\label{sec:intro}

Quantitative finance relies on licensed vendor feeds, index histories,
and alternative datasets. When a license expires, the licensee may be
required to stop using the data. Removing the stored data is not enough
if a model has already been trained on them. The model parameters may
still retain information learned from the expired data. The exact
remedy is to repeat the original training procedure without those data.
The retrained model is the counterfactual for deletion. Machine
unlearning seeks to reproduce this counterfactual without retraining.

Because repeated retraining is often costly, approximate methods are
commonly evaluated with proxy measures, whose reliability remains contested
\cite{nguyen2025survey,zhang2024mia_cannot}. The financial forecasters
studied here can instead be retrained in minutes. We call the model
trained on all available data the \emph{reference} and the model
retrained without the expired data the \emph{oracle}. We construct
3{,}200 paired comparisons on an S\&P~500 volatility-forecasting panel.
Within each pair, the reference and oracle follow the same training
procedure and forecast the same test days. Only the training data
differ. Three research questions organize the paper.

\textbf{RQ1: What does deleting a licensed era mean, and what does
exact deletion do?} A requested period can enter a forecasting pipeline
through targets, overlapping input windows, and later processing steps.
We define nested deletion levels to make the scope of the request
explicit. We then compare each reference with the oracle required by
that level. The main result concerns the 2020 COVID crisis year. It
produces the largest memorization gap for every architecture. After it
is removed, all three deployable models forecast better, with the
largest improvement in the 2022 bear market. Retaining valid crisis data
can therefore lower later forecast skill, even when the later market is
also volatile.

\textbf{RQ2: Can approximate unlearning substitute for retraining, and
what is the right target?} The correct target is the oracle, not an
inability to predict the deleted period. An oracle can still predict
that period by generalizing from retained data. Pushing predictions
toward noise instead moves away from the counterfactual and reduces
forecast skill. On this panel, method rankings do not transfer reliably
across architectures or rolling windows. Reliable use of approximate
unlearning therefore requires validation against the relevant oracle.

\textbf{RQ3: Can deletion be audited?} The answer depends on how much
the model memorized from the deleted period. When memorization is small,
models trained with and without that period can behave similarly even
after exact deletion. Window-level loss comparisons also provide limited
discrimination in this setting. A null audit therefore does not establish
that deletion succeeded. It may instead reflect insufficient detection
power.

\textbf{Contributions.} First, to our knowledge, we provide the first
systematic study of license-driven unlearning for a contiguous calendar
period in a rolling forecasting pipeline. We define a four-level
membership lattice for this request and quantify how its scope changes
both data cost and measured memorization. Second, we construct
3{,}200 paired reference--oracle comparisons across architectures,
rolling windows, deleted years, and deletion levels. This benchmark
measures how exact deletion changes both memorization and later forecast
skill. Third, we evaluate approximate methods and audits against the
same exact counterfactual. This evaluation distinguishes proximity to
the deletion counterfactual from loss of predictive ability on the
deleted period and shows that audit power depends on prior memorization.

Code, trained models, and complete results will be released upon
acceptance.

\section{Related Work}\label{sec:related}

Machine unlearning aims to remove the influence of requested training
data from a trained model~\cite{nguyen2025survey}. Exact approaches
prepare models for efficient deletion during
training~\cite{bourtoule2021sisa}, whereas approximate approaches edit
an existing model~\cite{kurmanji2023scrub}. Retraining without the
requested data defines the exact target but is often costly. For
temporal data, influence-based methods delete selected samples from load
forecasters~\cite{xu2024taskaware}. TS-Unlearn instead drives predictions
for randomly selected forget samples toward Gaussian
noise~\cite{wang2026tsunlearn}. Related work also discards the oldest
chunks of an image stream under concept drift~\cite{uil2026sliding}.
These studies define forget sets as selected samples, random subsets,
or stream chunks. They do not consider a calendar-period request in
which the same dates can enter multiple overlapping targets, input
windows, and derived features. They consequently do not compare nested
definitions of temporal membership or measure how the definition
changes the exact retraining counterfactual.

Unlearning audits often use membership inference to test whether
specified data were used for training. LiRA calibrates this decision
with shadow models~\cite{carlini2022lira}, and recent work extends such
attacks to time-series models~\cite{koren2025membership}. Without a
calibrated null population, however, membership inference cannot prove
that specified data were used~\cite{zhang2024mia_cannot}. Memorization
provides a second perspective: rare examples may need to be memorized
for i.i.d.\ generalization~\cite{feldman2020memorization}, and influence
scores quantify such memorization at the example
level~\cite{feldman2020what}. To our knowledge, prior work has not
studied license-driven unlearning of a contiguous calendar period in a
rolling forecasting pipeline. Our setting therefore requires both a
definition of temporal membership and evaluation against the
corresponding retraining oracle.

\section{Methodology}\label{sec:method}

\subsection{The unlearning problem}

Machine unlearning arises when part of a model's training data must
be deleted after training. Let $D$ be the training data and $A$ the
training procedure, including the architecture, the hyperparameters,
and the random seed. The \emph{forget set} $F \subset D$ holds the
observations to be deleted, and the \emph{retain set} $D \setminus F$
holds the rest. The model trained on everything, $A(D)$, is the
\emph{reference}. The model retrained from scratch on the retain
set, $A(D \setminus F)$, is the \emph{oracle}, or
\emph{counterfactual}, the ground truth of deletion. Because
retraining is often unaffordable, \emph{approximate unlearning}
instead edits the reference in place with an \emph{unlearning
operator} $U$, aiming for a model that a prescribed audit cannot
distinguish from the oracle.

\begin{table}[t]
\centering
\caption{Deletion levels for a requested period $T$. Each level
includes all restrictions of the preceding level.}
\label{tab:deletion_levels}
\small
\begin{tabular}{@{}p{0.08\columnwidth}p{0.85\columnwidth}@{}}
\toprule
Level & Prohibited use of $T$ and representative examples \\
\midrule
\Lzero{} &
\textbf{Target deletion.}
Use in forecast targets, such as prediction labels. \\
\Lone{} &
\textbf{Window deletion.}
Direct appearance in inputs or targets, such as context windows,
and use in the normalization statistics that scale them. \\
\Ltwo{} &
\textbf{Derived-data deletion.}
Dependence through feature construction, such as rolling statistics
and imputation. \\
\Lthree{} &
\textbf{Full-pipeline deletion.}
Influence on model-development decisions, such as hyperparameter
tuning and model selection. \\
\bottomrule
\end{tabular}
\end{table}

\subsection{A membership lattice for temporal deletion}

In time series, a deletion request does not name a forget set. It
names a period $T$, a set of dates whose use is no longer permitted.
A single date inside $T$ can touch a rolling forecaster in four
distinct ways, and each way is one level of
Table~\ref{tab:deletion_levels}. Consider one trading day in March 2020. Its return contributes to the
five-day targets of every window whose forecast horizon overlaps that
date (\Lzero). The same return can appear in the 64-day input contexts
of windows with later targets (\Lone). It may also affect derived
features or imputed values used by later windows (\Ltwo). Finally, it
may have influenced model-development decisions such as hyperparameter
tuning (\Lthree). Each level forbids everything the preceding level does
plus one further form of dependence, so the four readings are
nested. We call this ordering the membership lattice. The first three levels yield nested forget sets
$F_\ell(T)$ under the restrictions in Table~\ref{tab:levels}.

Strictness has two measurable consequences. The first is
data. Because windows overlap, \Lone{} removes windows whose target
dates lie outside $T$, and on this panel it removes about $25\%$
more windows than \Lzero. The second is the measurement itself. The
level changes the forget set, and with it the measured memorization
gap (Section~\ref{sec:levels}). A temporal deletion request is therefore incomplete unless its
level is named. We adopt \Lone{} as the paper-wide default and use
\Lzero{} and \Ltwo{} as robustness checks.
The lattice defines four conceptual levels, but only three yield
distinct experimental conditions. \Lthree{} coincides with \Ltwo{}
because all model-development decisions were made before the deletable
years.

At every level, the target of unlearning is the oracle retrained
without that level's forget set,
\begin{equation}
U\!\left(A(D),F_\ell(T)\right)
\;\approx\;
A\!\left(D\setminus F_\ell(T)\right),
\qquad
\ell\in\{0,1,2\},
\label{eq:unlearning}
\end{equation}
where $\approx$ is approximate equality in distribution under a
prescribed audit, and the method and the oracle use the same level.

\subsection{Counterfactuals and governance metrics}

For each deletion request and level, the oracle retrains the model from
scratch on the retain set under the reference model's training
protocol. An approximate method is judged by how closely it reproduces
this counterfactual, not by how much it degrades predictions on the
deleted period.

Let $\text{ref}=A(D)$ denote the reference model,
$\text{ora}=A(D\setminus F_\ell(T))$ the retrained counterfactual, and
$m=U(\text{ref},F_\ell(T))$ the model produced by an approximate
operator. The forget-set loss $\mathcal{L}_F(\cdot)$ measures prediction
error on the deleted data. A higher loss means worse fit, but not
necessarily more successful unlearning. Success is assessed relative to
$\mathcal{L}_F(\text{ora})$, not by maximizing $\mathcal{L}_F$. The test
score $s_{\text{test}}(\cdot)$ measures forecast quality on the test
set, oriented so that higher is better.

\textbf{Memorization gap $G_F$.} The memorization gap is the reference
model's excess fit over the counterfactual on the forget set,
\begin{equation}
G_F = \mathcal{L}_F(\text{ora}) - \mathcal{L}_F(\text{ref}).
\label{eq:gap}
\end{equation}
A gap near zero means that the forget set was barely memorized, while a
large positive gap indicates stronger memorization. Because
$\mathcal{L}_F$ is an MSE, $G_F$ has squared target units. For
comparisons across models trained on differently scaled targets, we
report the dimensionless gap
$G_F/\operatorname{Var}(y_T)$, where $y_T$ denotes the targets in the
deleted period.

\textbf{Closure $\gapclosed$.} When $G_F>0$, closure measures how far a
method moves from the reference loss toward the oracle loss:
\begin{equation}
\gapclosed(m) =
\frac{\mathcal{L}_F(m)-\mathcal{L}_F(\text{ref})}{G_F}.
\label{eq:kappa}
\end{equation}
A closure of $0$ matches the reference loss, while a closure of $1$
matches the oracle loss. Values above $1$ indicate that the method has
higher forget-set loss than the oracle. When $G_F$ is near zero, the
ratio is unstable and closure is not interpretable.

\textbf{Deletion effect $\Delta s_{\text{test}}$.} The deletion effect
is the change in test score caused by exact deletion, measured one day
at a time so that offsetting daily differences are not hidden by the
average. The reference and the counterfactual score the same
cross-section on each evaluation day $t$, so the daily change is
$\Delta s_t = s_t(\text{ora}) - s_t(\text{ref})$, and
\begin{equation}
\Delta s_{\text{test}} =
\frac{1}{N_{\text{test}}}\sum_{t=1}^{N_{\text{test}}} \Delta s_t .
\label{eq:delta}
\end{equation}
A positive $\Delta s_{\text{test}}$ means deletion improves the
test score, and a negative value means the deleted period carried
useful signal. Where no confusion arises we write $\Delta$ for
$\Delta s_{\text{test}}$ and $G$ for $G_F$.

\subsection{Auditing deletion}

The goal of an audit is to determine whether information from the
deleted period remains detectable in the model. 
Throughout,
the audit forms two populations by retraining under different random
seeds: references that include $T$ in training, and counterfactuals
that omit it. This is the access model a model owner or an appointed
auditor can realistically arrange: fixed code, fixed data cuts, and
multiple retraining seeds.
At the era level, the audit statistic is the forget-set loss
$\mathcal{L}_F$. Across training seeds, the audit separation is
\begin{equation}
d' =
\frac{
\overline{\mathcal{L}_F(\text{ora})}
-
\overline{\mathcal{L}_F(\text{ref})}
}{
\mathrm{sd}_{\text{pool}}
},
\label{eq:dprime}
\end{equation}
where the bars denote means across seeds and
$\mathrm{sd}_{\text{pool}}$ is the pooled standard deviation of the two
populations. A positive $d'$ means that the oracles have higher
forget-set loss than the references, as expected after deletion.
Larger positive values indicate stronger separation, while
$d'\approx 0$ means that the populations overlap. A negative $d'$
means that the observed difference runs in the opposite direction.

Audit granularity also matters. The era-level audit uses the average
forget-set loss over the deleted period. At the window level, we report
the fraction of deleted windows for which the reference has lower loss
than its paired oracle; a value of $1/2$ indicates chance.

\section{Experimental Setup}\label{sec:exp}

\subsection{One run of the study}

We train a Transformer on the five years 2016--2020 and deploy it on
2022. Now the license on the 2020 data expires. We retrain the same
architecture, with the same hyperparameters and the same seed, on the
remaining four years. The original model is the reference of
Section~\ref{sec:method}, and the retrained model is the oracle. For one
seed, this pair defines the memorization gap $G$ on 2020 and the
deletion effect $\Delta$ on 2022. Repeating the comparison across seeds
provides the reference and oracle populations used to audit deletion.

The full study contains 320 cells with ten seeds each. The main grid
has 300 cells: five architectures, four folds, five deletable years,
and three deletion levels. Twenty random-deletion placebo cells bring
the total to 3{,}200 paired comparisons.

\subsection{Forecasting task and data}

The task is 5-day-ahead realized volatility for S\&P~500 stocks, one
cross-section per trading day: each day the model ranks stocks by how
volatile the next five days will be. This is a risk task rather than
an alpha contest; the ranking feeds volatility targeting and position
sizing. For each asset $i$ and day $t$ the model reads a 64-day
history of eleven return-derived features and predicts
$y^{\mathrm{raw}}_{i,t} = \log
\operatorname{sd}\{r_{i,t+1},\ldots,r_{i,t+5}\}$, the log
volatility of its next five daily returns, $z$-scored within each
day's cross-section and clipped to $[-3,3]$. Training minimizes
mean squared error against this target.

Daily returns are obtained from Yahoo Finance and adjusted for
dividends and splits. We use a fixed list of 501 current S\&P~500
constituents, so delisted firms are excluded by construction. The panel
covers 2015--2025 and contains about 480 stocks per day.
Hyperparameter tuning uses an earlier extract reaching back to 2008.

\subsection{Rolling windows and deletions}

We use a five-year rolling training window slid
forward one year at a time. Each position of the window validates
on the following year and tests on the year after that. We call
each window position a \emph{fold}, named by its test year: the
2022 fold trains on 2016--2020, validates on 2021, and tests on
2022, and the folds run through 2025. Rolling fixed-width windows
are standard deployment practice and give every era a scheduled
exit, which makes deleting a year still inside the active window
the live compliance case.

Sliding the window allows two comparisons. Within a fold, varying the
deleted year compares different years under the same training window
and test market. Across folds, repeatedly deleting 2020 shows how its
effect changes with the test market, the training window, and the
position of 2020 within that window.

A random-deletion placebo completes the design. It removes the same
number of windows as the \Lone{} deletion of 2020, sampled uniformly
from the training set. This separates the effect of losing data volume
from the effect of losing that particular year. We do not top up the
training set because the compliance counterfactual after license expiry
contains less training data.

\subsection{Models and benchmark}

We tune seven forecasting backbones and report five spanning the
observed fit range: the vanilla
Transformer~\cite{vaswani2017attention},
TSMixer~\cite{chen2023tsmixer}, SegRNN~\cite{lin2023segrnn},
iTransformer~\cite{liu2024itransformer}, and
PatchTST~\cite{nie2023patchtst}. The other two will be reported in the
released sweep.

As an econometric benchmark, we refit the heterogeneous autoregressive
(HAR) model of realized volatility~\cite{corsi2009har} under the same
rolling-window and deletion protocol. It regresses log 5-day-ahead
realized volatility on its trailing 1-, 5-, and 22-day counterparts by
ordinary least squares and has four parameters. We classify an
architecture as deployable if its test IC exceeds that of HAR in every
fold. Three architectures meet this criterion and are the
\emph{deployable} models. The other two never outperform HAR, underfit
this task, and are the \emph{non-deployable} models.

\subsection{Metrics, tuning, and cost}

Forecast skill is the daily cross-sectional information coefficient
(IC): the Spearman rank correlation between predictions and targets
across stocks, averaged over the test year's trading days. We write this
score as $s_{\text{test}}$. The corresponding IC on windows whose
five-day forecast targets overlap the requested year $T$ is written
\forgetic{}. The forget-set loss $\mathcal{L}_F$ is the mean squared
error over the level-specific forget set $F_\ell(T)$. The metrics $G$, $\Delta$, and $\gapclosed$ are defined in
Eqs.~\ref{eq:gap}--\ref{eq:delta}. Test-score differences are paired by
trading day and use Newey--West standard
errors~\cite{newey1987hac}. Forget-set metrics are averaged over ten
seeds. The audit separation $d'$ compares the corresponding ten-seed
reference and oracle populations.

Hyperparameters were tuned once, on a split that predates every
deletable year (2010--2014, validated on 2015, disjoint seeds),
selecting on validation loss over a grid of width and dropout. An expanded search over learning rates and structural settings did not
move iTransformer or PatchTST above HAR. A single oracle retrain costs 3--18 minutes on
one NVIDIA A100, and the full campaign took about 600 A100-hours.

\section{RQ1: What Does Exact Deletion Do?}\label{sec:rq1}

This section compares the reference with the oracle after one training
year is removed. The memorization gap $G$ (Eq.~\ref{eq:gap}) is
evaluated on the level-specific forget set, while the deletion effect
$\Delta$ (Eq.~\ref{eq:delta}) is evaluated on the later test year.
Positive values indicate, respectively, greater memorization and
improved forecast skill after deletion. We examine how both vary across
deleted years, folds, and deletion levels.



Tables~\ref{tab:years} and~\ref{tab:skill} provide the market and model
context. Table~\ref{tab:years} shows that the 2020 crisis year has
$38\%$ realized volatility, compared with $24\%$ in the 2022 bear
market and $7$--$16\%$ in the remaining deletable years.
Table~\ref{tab:skill} reports reference skill. The Transformer, TSMixer,
and SegRNN exceed HAR in every fold and are the deployable models.
iTransformer and PatchTST never exceed HAR and are the non-deployable
models.

\begin{table}[t]
\caption{Annualized realized volatility of the deletable years. Each
year appears in at least one training window.}
\label{tab:years}
\centering
\setlength{\tabcolsep}{3.6pt}
\small
\begin{tabular}{lcccccccc}
\toprule
Year & 2016 & 2017 & 2018 & 2019 & \textbf{2020} & 2021 & \textbf{2022} & 2023\\
\midrule
Realized vol.\ (\%) & 14 & 7 & 16 & 13 & $\mathbf{38}$ & 14 & $\mathbf{24}$ & 14\\
\bottomrule
\end{tabular}
\end{table}

\begin{table}[t]
\caption{Reference skill (IC) per fold. Values are means across ten seeds. Rows are sorted by skill,
and the refitted HAR benchmark, evaluated on the same test years,
separates the deployable models above it from the non-deployable
models below it. Folds are ordered by test-year volatility.}
\label{tab:skill}
\centering
\setlength{\tabcolsep}{5pt}
\small
\begin{tabular}{lcccc}
\toprule
& 2022 & 2025 & 2023 & 2024\\
Architecture & ($24\%$) & ($18\%$) & ($14\%$) & ($12\%$)\\
\midrule
Transformer & $0.570$ & $0.454$ & $0.470$ & $0.489$\\
TSMixer & $0.560$ & $0.463$ & $0.458$ & $0.488$\\
SegRNN & $0.517$ & $0.452$ & $0.441$ & $0.476$\\
\midrule
HAR (benchmark) & $0.508$ & $0.412$ & $0.384$ & $0.396$\\
\midrule
iTransformer & $0.125$ & $0.213$ & $0.174$ & $0.221$\\
PatchTST & $0.168$ & $0.181$ & $0.181$ & $0.169$\\
\bottomrule
\end{tabular}
\end{table}

\begin{figure}[t]
\centering
\includegraphics[width=\columnwidth]{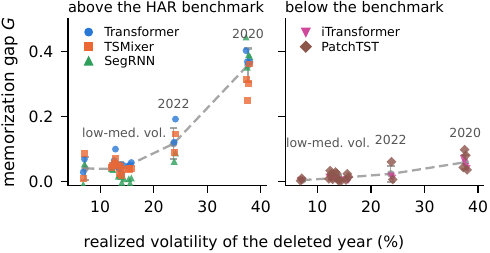}
\caption{Memorization gap $G$ for all 100 cells (5 architectures
$\times$ 4 folds $\times$ 5 deleted years, \Lone; each point the
mean of 10 seeds), placed at the realized volatility of the deleted
year and jittered horizontally by fold. Left: the three deployable
architectures. Right: the two non-deployable ones, on the same
vertical scale. The grey dashed curve connects the means of the
four volatility clusters, bars $\pm 1$ sd.}
\label{fig:gap}
\end{figure}

\subsection{Memorization of training years}

Figure~\ref{fig:gap} shows the memorization gap for every combination
of architecture, fold, and deleted year. Each point is placed at the
realized volatility of the deleted year. The gap increases with
volatility for every architecture. Deleting 2020 produces the largest
gap in every fold, the low- to medium-volatility years produce gaps
near zero, and the 2022 bear market lies between the two. Models therefore memorize
volatile years far more strongly than low- to medium-volatility ones.

The size of the gap also differs across architectures. For every
deleted year, the three deployable architectures in the left panel
have larger gaps than the two non-deployable architectures in the
right panel. HAR shows the same volatility pattern at a smaller
magnitude. Memorization is therefore associated with both the
volatility of the deleted year and the architecture.

A size-matched random deletion produces a gap near zero, so the number
of removed windows does not explain the pattern. The gap is also
unrelated to the reference model's fit to each year, so year-level
difficulty does not explain it. These results are consistent with the
crisis year having fewer substitutes in the retained data, but they do
not identify the mechanism.

\begin{figure}[t]
\centering
\includegraphics[width=\columnwidth]{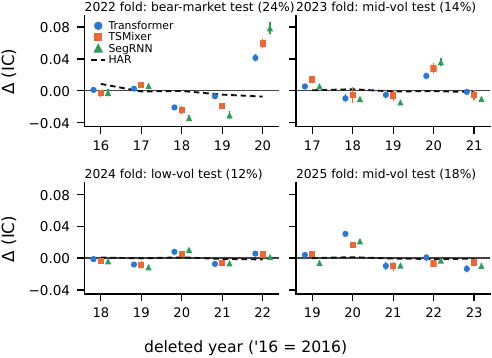}
\caption{Deletion effect $\Delta$ from removing each training year in
every fold (\Lone). Each point is the mean across 10 seeds. All panels
share the same scale, and points are offset horizontally by
architecture for visibility. The dashed black line shows the deletion
effect of the refitted HAR model under the same year deletion. Only the
deployable models are shown.}

\label{fig:ladder}
\end{figure}

\subsection{Effect of deletion on forecast skill}

Figure~\ref{fig:ladder} reports the deletion effect for the three
deployable architectures. Each panel represents one fold, and each
point gives the change in test skill after one training year is
removed. A positive value means that the model forecasts better
without that year. Deleting 2020 improves all three architectures in
all four folds. The improvement is largest in the 2022 bear market and
smaller in the later low- to medium-volatility test years. In the 2024
fold, no architecture changes by more than about $3\%$ of its skill,
which is close to the placebo effect. The crisis year is therefore the
only deleted year whose removal consistently improves forecasts.

If the 2020 observations were intrinsically harmful, every model should
improve when they are removed. The HAR model instead loses a small
amount of skill and has a much smaller memorization gap. The 2020 data
therefore remain useful to HAR, while their removal benefits the
deployable models. The improvement is specific to these architectures,
not an inherent property of the data. Prior work shows that memorizing
atypical examples can improve generalization under i.i.d.\
evaluation~\cite{feldman2020memorization}. Here, the test years are
later temporal regimes rather than i.i.d.\ draws from the training
distribution. Removing the crisis year improves all three deployable
architectures, but the data do not identify which model property causes
this result. Test skill and memorization increase together across the
five architectures.

The two non-deployable architectures show no comparable
regularity. Their deletion effects change sign across folds and
deletion levels, and PatchTST loses about $40\%$ of its skill in the
bear-market fold. By contrast, the deployable models respond to deletion
consistently across folds and deletion levels. Their gains are also not
carried by a few turbulent days. The oracle
outperforms the reference on the majority of individual test days.

\subsection{Dependence on year and recency}

Comparing the five deletions within each panel of
Figure~\ref{fig:ladder} shows two patterns. Removing an early
low- to medium-volatility year has little effect. Removing a recent
low- to medium-volatility year generally reduces skill. In contrast,
removing the crisis year improves skill at every position it occupies
in the rolling window.

Year identity and position vary together in these comparisons, so the
design does not isolate a separate causal effect of recency. The
results nevertheless show that neither volatility nor recency alone
determines the deletion effect. Recent low- to medium-volatility years
contain useful information, while the crisis year can reduce later
skill for deployable models. We do not identify the mechanism. The retraining oracle measures the
net effect directly.

\subsection{Dependence on the deletion level}\label{sec:levels}

\begin{table}[t]
\caption{Deletion scope and measurements for 2020. Forget-set size is
reported as a share of all training windows for \Lzero{} and as the
increase over the preceding level for \Lone{} and \Ltwo{}.
$G(2020)$ is the Transformer memorization gap averaged over folds and
seeds. The final column reports the number of deployable cells with a
positive deletion effect. Parentheses report how many are statistically
significant.}
\label{tab:levels}
\centering
\small
\begin{tabular}{llcc}
\toprule
Level & Forget-set size & $G(2020)$ & Positive cells \\
\midrule
\Lzero & $20\%$ of all windows & $0.527$ &
$12/12$ (all significant) \\
\Lone & $25\%$ more than \Lzero & $0.375$ &
$12/12$ (all significant) \\
\Ltwo & $2\%$ more than \Lone & $0.368$ &
$12/12$ ($11/12$ significant) \\
\bottomrule
\end{tabular}
\end{table}

All results so far use \Lone{}. The grid
repeats each deletion at \Lzero{} and \Ltwo{} with the same ten
seeds, so the data cost of stricter deletion is measured rather than assumed.
Table~\ref{tab:levels} summarizes what the level changes.

The crisis-year gap falls at each stricter deletion level in all twenty
architecture-fold cells. Nearly all of the decline occurs from
\Lzero{} to \Lone{}, which adds context-overlapping windows and
recomputes normalization. Moving from \Lone{} to \Ltwo{} changes the
gap only slightly. The gaps for the low- to medium-volatility years
change less, and their directions vary by year. Two contracts naming
the same period at different levels can therefore measure different
amounts of memorization.

Deleting 2020 improves every deployable architecture at all three
deletion levels. The losses from deleting recent low- to
medium-volatility years are also present at \Lzero{} in the bear-market
fold, so they are not caused by the wider \Lone{} forget set. In
contrast, the effects for the non-deployable models vary across levels;
for example, iTransformer's positive delete-2020 effect at \Lzero{}
becomes negligible at \Lone{}. The deletion level changes data cost and measured memorization, but not
the direction of the findings for the deployable models. RQ2 asks
whether retraining can be avoided once the deletion level is fixed.

\section{RQ2: Can Retraining Be Avoided, and What Is the Target?}
\label{sec:rq2}

An approximate unlearning method edits the reference model in place
instead of retraining it. This section asks whether an edited model
can substitute for the oracle. We first define the target and then
evaluate four methods against it.

Closeness is measured by the closure $\gapclosed$
(Eq.~\ref{eq:kappa}), evaluated on the level-specific forget set. When
$G>0$, a closure of $0$ matches the reference loss, a closure of $1$
matches the oracle loss, and values above $1$ exceed the oracle loss.
All method results use the crisis-year deletion. For other years, the
oracle's memorization gap is too close to zero for closure to be
interpretable.

We test four methods. Finetune continues training on the retained data.
The hinge baseline minimizes
\begin{equation}
\mathcal{L}_{\mathrm{hinge}}
=
\mathcal{L}_R
+
\alpha\max\{0,\tau-\mathcal{L}_F\},
\label{eq:hinge}
\end{equation}
where $\mathcal{L}_R$ and $\mathcal{L}_F$ are minibatch mean squared
errors on the retain and forget sets. We fix $\alpha=0.5$ and $\tau=1$
before evaluation. Because the targets are cross-sectionally
standardized each day, a zero predictor has an MSE of approximately
one. The hinge penalizes forget-set losses below this fixed threshold
while continuing to minimize retain-set loss. SCRUB distills the
reference on the retained data and diverges
from it on the forget set~\cite{kurmanji2023scrub}. TS-Unlearn, which we
reimplement, is designed specifically for time-series
forecasting~\cite{wang2026tsunlearn}. All remaining hyperparameters are
prespecified and will be reported in the released protocol. Plain
gradient ascent diverges in every cell and is excluded.

\subsection{The target of forgetting}

\begin{figure}[t]
\centering
\includegraphics[width=\columnwidth]{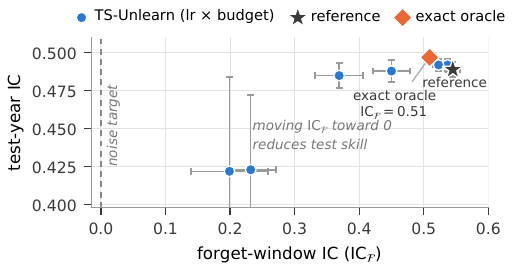}
\caption{The oracle and noise targets, compared on the Transformer
under the crisis-year deletion in the 2024 fold (\Lone). Each point
is one TS-Unlearn configuration (learning rate $\times$ budget),
averaged over ten seeds, with bars showing $\pm 1$ sd across seeds. The
dashed line marks the noise target, the diamond the exact
counterfactual, and the star the reference. The counterfactual has
higher test skill than the reference because deleting the crisis year
helps this model.}
\label{fig:twodefs}
\end{figure}

The methods compared here imply two targets. The oracle target asks the
unlearned model $m$ to behave like the retrained counterfactual. The
noise target used by TS-Unlearn instead drives predictions on the
forget set toward Gaussian noise. With $X_f$ denoting the forget-set
inputs,
\begin{equation}
\text{oracle: } m \approx_d \text{ora},
\qquad
\text{noise: } m(X_f) \stackrel{d}{\to} \mathcal{N}(0,\Sigma).
\label{eq:compliance}
\end{equation}
Figure~\ref{fig:twodefs} compares the two targets in one picture.
We take the reference Transformer in the 2024 fold, delete the
crisis year, and apply TS-Unlearn at increasingly aggressive
settings, which produces a
family of unlearned models. Every model in the figure is placed by two
numbers. The horizontal axis is its forget-set skill, $\forgetic$, and
the vertical axis is its test-year skill. The reference model is the
star at the upper right and predicts both sets well. The noise target
is the dashed line at zero, where forget-set predictions are no better
than noise. The oracle target is the diamond, the exact counterfactual
retrained without the year.

The diamond lies near the reference rather than the noise line. The
counterfactual never trained on 2020, yet it predicts 2020 at
$\forgetic = 0.51$, barely below the reference's $0.55$. Most skill on
the removed year is therefore generalization from the retained years,
not memorization of the year itself. A model that predicts the removed
year like noise does not match the retrained counterfactual.

The blue points show the effect of moving toward the noise target. As
the settings become more aggressive, forget-set skill approaches zero
while test skill falls from $0.49$ to $0.42$. The error bars widen
about tenfold, so the models closest to the noise target are also the
least stable across seeds. This is the 2024 fold, where exact deletion
changes skill by at most $3\%$. The noise target therefore reduces test
skill even when exact deletion has little effect. The appropriate
target for license-driven deletion is the counterfactual.

\begin{figure}[t]
\centering
\includegraphics[width=\columnwidth]{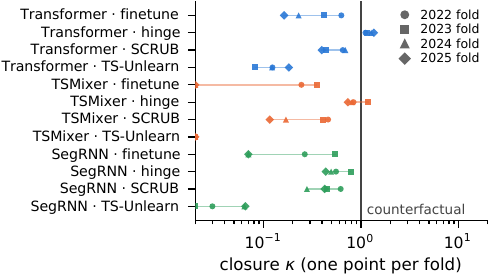}
\caption{Closure $\gapclosed$ for each deployable
architecture--method pair on the crisis-year deletion (\Lone). Each
marker is the mean across ten seeds at fixed, prespecified
hyperparameters. Each fold contributes one marker, and a thin line
connects the four folds. A pair substitutes for the oracle only when
its markers remain near the counterfactual line. Values at or below
zero are clipped to the left edge of the log axis.}
\label{fig:methods}
\end{figure}

\subsection{Substitution for the oracle}

Figure~\ref{fig:methods} evaluates the four methods on the three
deployable architectures. The two non-deployable ones are omitted
from the figure. Each row is one architecture-method pair, with one
marker per fold. The test year plays no part in the closure. What
changes across folds is the training window and the position of
2020 inside it. A tightly clustered row indicates that a method tracks
the oracle across window positions. A dispersed row indicates that its
result depends on the window in which it is applied.

The vertical line at $\gapclosed=1$ marks the counterfactual. In any
single fold, several pairs lie near this line. Across all four folds,
only the TSMixer--hinge pair remains near it. Its closure ranges from
$0.74$ to $1.18$, with no measurable loss of test skill.

The hinge does not estimate the oracle target. For the TSMixer
crisis-year deletions, the fixed threshold $\tau=1$ happens to lie
between the reference and oracle forget-set losses. It therefore stops
near the counterfactual in these cells. This alignment is specific to
the architecture and deleted era, rather than an oracle estimate
produced by the method. For the low- to medium-volatility years, the
oracle forget-set loss lies below $\tau$, so the same threshold would
overshoot the counterfactual.

The Transformer--hinge pair is second best. It overshoots the oracle in
every fold but remains within half of the memorization gap. No other
pair stays within half of the gap in every fold. Most failures of
Finetune and SCRUB are due to insufficient movement from the reference,
whereas the direction of the hinge error depends on the architecture.

TS-Unlearn fails under both the prespecified budget and more aggressive
settings. At the prespecified budget in Figure~\ref{fig:methods}, its
closure is $0.2$ or less on every architecture. This budget also
reduces skill. In the bear-market fold, test skill falls
$0.07$--$0.13$ IC short of the oracle and below the reference. At the
more aggressive settings in Figure~\ref{fig:twodefs}, TS-Unlearn
overshoots the counterfactual and loses further skill.

The rankings do not transfer across architectures. The same hinge
under-forgets on SegRNN, brackets the counterfactual on TSMixer,
and overshoots on the Transformer. 
Their
failures are more severe. The hinge overshoots about fivefold on
iTransformer, and SCRUB spans two orders of magnitude across seeds on
PatchTST. For PatchTST, selecting the method nearest the
counterfactual is misleading because the mechanical winner barely
moves from the reference. The memorization gap does not predict which
method will perform best for an architecture.

Fold-to-fold variation also has no consistent pattern. The ordering of
the folds changes across architecture--method pairs. The oracle gap for
the 2020 deletion varies by less than half across folds on the
deployable architectures, while method closures cross from one side of
the counterfactual to the other. Variation in the oracle gap therefore
does not explain the method variation. Table~\ref{tab:method_choice}
lists the method that minimizes the worst-fold absolute deviation from
$\gapclosed=1$ for each deployable architecture.

None of the approximate methods studied here eliminates the need for
oracle validation. A closure of $1$ is defined by the oracle's loss on
the level-specific forget set, so a model owner cannot locate the exact
target without retraining. The TSMixer--hinge pair is the only
combination that remains near the oracle across all four folds. Its
success reflects an empirical alignment between the fixed threshold
and the crisis-year oracle loss, not an ability to locate the
counterfactual without retraining. Results do not transfer across
architectures or window positions, and the memorization gap does not
predict method performance. In these experiments, reliable acceptance
therefore requires validation against an oracle for the same
architecture and window. This validation must be repeated as the
window slides, and each retraining run costs minutes.

\begin{table}[t]
\caption{The best method per deployable architecture, selected to
minimize the maximum $|\gapclosed-1|$ across folds. The closure range
spans the four folds. Skill cost is the worst-fold decrease in test IC
relative to the paired oracle, and the last column is wall-clock
cost relative to retraining.}
\label{tab:method_choice}
\centering
\small
\begin{tabular}{llccc}
\toprule
Architecture & Best method & Closure $\gapclosed$ & Skill cost & Cost \\
\midrule
TSMixer & hinge & $0.74$--$1.18$ & $0.004$ & $0.4\times$ \\
Transformer & hinge & $1.12$--$1.36$ & $0.011$ & $0.4\times$ \\
SegRNN & hinge & $0.44$--$0.79$ & $0.004$ & $0.2\times$ \\
\bottomrule
\end{tabular}
\end{table}

\section{RQ3: Can Deletion Be Audited?}\label{sec:rq3}

A loss-based audit can distinguish deletion only if models trained
with and without the requested era behave differently. We estimate
these two populations using ten references and ten paired oracles and
measure their standardized separation by $d'$
(Eq.~\ref{eq:dprime}). A value near zero means that the populations
overlap, so this statistic provides little evidence for distinguishing
them. We examine separation at the level of an era and an individual
training window, then measure how treating overlapping windows as
independent affects statistical confidence.

\begin{figure}[t]
\centering
\includegraphics[width=\columnwidth]{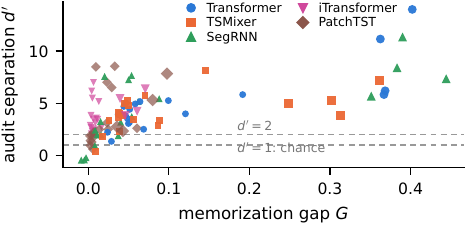}
\caption{Audit separation $d'$ against the memorization gap $G$
for all 100 cells of architecture, deleted year, and fold (\Lone).
Crisis-year points are drawn larger, and the dashed lines mark
$d' = 1$ and $d' = 2$ as descriptive reference levels. All five
architectures are included because the applicability of an audit does
not depend on whether a model is deployable.}
\label{fig:audit}
\end{figure}

\subsection{Era-level auditing}

Figure~\ref{fig:audit} reports the audit in all 100 cells of
architecture, deleted year, and fold. Each point places one cell by its
memorization gap from Section~\ref{sec:rq1} on the horizontal axis and
its audit separation on the vertical axis. The figure shows how
population separation varies with prior memorization.

Across the grid the separation spans $-0.4$ to $14.0$. Removing the
crisis year from a model with a large memorization gap strongly
separates the two populations. Removing a low- to medium-volatility
year with a near-zero gap leaves substantial overlap. Separation rises
with the gap across the grid, and the cells with the smallest values of
$d'$ also have small memorization gaps. Seventeen per cent of the cells
fall below the descriptive reference level $d'=2$. This reference level
does not by itself define a pass--fail rule for deletion.

Exact deletion can therefore produce little observable change when the
reference memorized little from the requested era. In these cells, the
oracle construction confirms deletion even though the loss
distributions overlap. A null result is therefore inconclusive: it can
reflect successful deletion with little prior memorization or a failure
to delete. Audit results must be interpreted together with their
detection power.

\subsection{Window-level auditing}

We next test the finest granularity available in this pipeline by
scoring each deleted window separately. For every window and paired
seed, we record whether the reference has lower squared error than its
oracle and average this indicator within each cell. The chance level of
this win rate is one half. Its median is just above one half, and the
largest value in the grid is $0.69$. The win rate rises with
memorization, as the era-level separation does, but provides limited
discrimination at the level of an individual window.

\subsection{Overlapping windows and audit power}

The window count overstates the audit's evidence unless this dependence
is respected. Each forecast date contributes about 480 stock-level
windows that share one market cross-section, while adjacent dates have
overlapping inputs and forecast horizons. We therefore aggregate losses
by date and account for serial dependence with HAC standard errors.
Relative to this calculation, treating all stock-level windows as
independent inflates the absolute $t$-statistic by a median factor of
$1.9$ across the grid.

After aggregation by date, the remaining uncertainty includes
variation across training seeds. Adding more correlated windows from
the same trained models does not remove this variation; additional
independent retraining seeds are required to estimate population
separation more precisely. Audit reports should therefore state the
aggregation level, the number of retraining seeds, and the detectable
effect size alongside $d'$.

\section{Conclusion}\label{sec:conclusion}

Exact retraining provides a direct standard for evaluating temporal
deletion. On this panel, removing the crisis year improves every
deployable architecture across four folds. The effect depends on the
architecture, the test regime, and the deleted year's position in the
training window. Approximate methods rarely match the oracle, and
their rankings do not transfer across architectures or folds. The
counterfactual remains predictive on the deleted year, so forcing
predictions toward noise is not a valid compliance target. Audit
strength depends on how much the model memorized, and a null audit does
not establish compliance.

\textbf{Practice.} A data contract should specify the deletion level,
the comparison standard, and the audit granularity. The appropriate
standard is equivalence to the retrained counterfactual, not inability
to predict the deleted period. Era-level audits can provide evidence
of deletion when memorization is sufficient. The window-level statistic
studied here provides limited discrimination between references and
oracles.

\textbf{Limitations.} The evidence comes from one liquid US equity
panel, one forecasting task, and five architectures. Test skill and
memorization are collinear across these architectures, so the data do
not identify which model property explains the deletion effect. A
single crisis year also cannot establish why removing 2020 improves
later forecasts. All experiments use fixed five-year windows and one
configuration per architecture. Other assets, window designs, and
selection rules remain to be tested. The protocol, configurations, per-cell artifacts, and verification
scripts will be released upon acceptance.

\bibliographystyle{ACM-Reference-Format}
\bibliography{refs}

@inproceedings{wang2026tsunlearn,
  title={TS-Unlearn: A Dual-Objective Unlearning Framework for Time Series Forecasting},
  author={Wang, Ziyi and Chen, Lixing and Miao, Zhongqi and Tang, Junhua and Bai, Yang and Li, Jianhua},
  booktitle={Pacific-Asia Conference on Knowledge Discovery and Data Mining},
  pages={341--353},
  year={2026},
  organization={Springer}
}

@article{nguyen2025survey,
  author  = {Nguyen, Thanh Tam and Huynh, Thanh Trung and Ren, Zhao and Nguyen, Phi Le and Liew, Alan Wee-Chung and Yin, Hongzhi and Nguyen, Quoc Viet Hung},
  title   = {A Survey of Machine Unlearning},
  journal = {ACM Transactions on Intelligent Systems and Technology},
  volume  = {16},
  number  = {5},
  articleno = {108},
  pages   = {1--46},
  year    = {2025},
  doi     = {10.1145/3749987}
}

@inproceedings{bourtoule2021sisa,
  author    = {Bourtoule, Lucas and Chandrasekaran, Varun and Choquette-Choo, Christopher A. and Jia, Hengrui and Travers, Adelin and Zhang, Baiwu and Lie, David and Papernot, Nicolas},
  title     = {Machine Unlearning},
  booktitle = {IEEE Symposium on Security and Privacy (S\&P)},
  pages     = {141--159},
  year      = {2021},
  doi       = {10.1109/SP40001.2021.00019}
}

@inproceedings{kurmanji2023scrub,
  author    = {Kurmanji, Meghdad and Triantafillou, Peter and Hayes, Jamie and Triantafillou, Eleni},
  title     = {Towards Unbounded Machine Unlearning},
  booktitle = {Advances in Neural Information Processing Systems (NeurIPS)},
  volume    = {36},
  pages     = {1957--1987},
  year      = {2023}
}

@inproceedings{carlini2022lira,
  author    = {Carlini, Nicholas and Chien, Steve and Nasr, Milad and Song, Shuang and Terzis, Andreas and Tram{\`e}r, Florian},
  title     = {Membership Inference Attacks From First Principles},
  booktitle = {IEEE Symposium on Security and Privacy (S\&P)},
  pages     = {1897--1914},
  year      = {2022},
  doi       = {10.1109/SP46214.2022.9833649}
}

@article{zhang2024mia_cannot,
  author  = {Zhang, Jie and Das, Debeshee and Kamath, Gautam and Tram{\`e}r, Florian},
  title   = {Membership Inference Attacks Cannot Prove that a Model Was Trained on Your Data},
  journal = {arXiv preprint arXiv:2409.19798},
  year    = {2024}
}

@article{uil2026sliding,
  author  = {Micha{\l} Wo{\'z}niak and Marek Klonowski and Maciej M{\k{a}}czy{\'n}ski and Bartosz Krawczyk},
  title   = {Unlearning-Based Sliding Window for Continual Learning under Concept Drift},
  journal = {arXiv preprint arXiv:2603.14484},
  year    = {2026}
}

@article{newey1987hac,
  author  = {Newey, Whitney K. and West, Kenneth D.},
  title   = {A Simple, Positive Semi-definite, Heteroskedasticity and Autocorrelation Consistent Covariance Matrix},
  journal = {Econometrica},
  volume  = {55},
  number  = {3},
  pages   = {703--708},
  year    = {1987}
}

@inproceedings{feldman2020memorization,
  author = {Vitaly Feldman},
  title = {Does Learning Require Memorization? {A} Short Tale about a Long Tail},
  booktitle = {Proceedings of the 52nd Annual ACM SIGACT Symposium on Theory of Computing (STOC)},
  pages = {954--959},
  year = {2020},
}

@inproceedings{feldman2020what,
  author = {Vitaly Feldman and Chiyuan Zhang},
  title = {What Neural Networks Memorize and Why: Discovering the Long Tail via Influence Estimation},
  booktitle = {Advances in Neural Information Processing Systems (NeurIPS)},
  volume = {33},
  year = {2020},
}

@article{vaswani2017attention,
  title={Attention is all you need},
  author={Vaswani, Ashish and Shazeer, Noam and Parmar, Niki and Uszkoreit, Jakob and Jones, Llion and Gomez, Aidan N and Kaiser, {\L}ukasz and Polosukhin, Illia},
  journal={Advances in neural information processing systems},
  volume={30},
  year={2017}
}

@inproceedings{nie2023patchtst,
  author    = {Nie, Yuqi and Nguyen, Nam H. and Sinthong, Phanwadee and
               Kalagnanam, Jayant},
  title     = {A Time Series is Worth 64 Words: Long-term Forecasting with
               Transformers},
  booktitle = {International Conference on Learning Representations (ICLR)},
  year      = {2023}
}

@inproceedings{liu2024itransformer,
  title={itransformer: Inverted transformers are effective for time series forecasting},
  author={Liu, Yong and Hu, Tengge and Zhang, Haoran and Wu, Haixu and Wang, Shiyu and Ma, Lintao and Long, Mingsheng},
  booktitle={International conference on learning representations},
  volume={2024},
  pages={11116--11140},
  year={2024}
}

@article{lin2023segrnn,
  title={Segrnn: Segment recurrent neural network for long-term time series forecasting},
  author={Lin, Shengsheng and Lin, Weiwei and Wu, Wentai and Zhao, Feiyu and Mo, Ruichao and Zhang, Haotong},
  journal={IEEE Internet of Things Journal},
  year={2025},
  publisher={IEEE}
}

@article{chen2023tsmixer,
  author  = {Chen, Si-An and Li, Chun-Liang and Yoder, Nate and Arik, Sercan~O.
             and Pfister, Tomas},
  title   = {{TSMixer}: An All-MLP Architecture for Time Series Forecasting},
  journal = {Transactions on Machine Learning Research (TMLR)},
  year    = {2023}
}

@article{corsi2009har,
  author  = {Corsi, Fulvio},
  title   = {A Simple Approximate Long-Memory Model of Realized Volatility},
  journal = {Journal of Financial Econometrics},
  volume  = {7},
  number  = {2},
  pages   = {174--196},
  year    = {2009},
  doi     = {10.1093/jjfinec/nbp001}
}

@article{xu2024taskaware,
  author  = {Xu, Wangkun and Teng, Fei},
  title   = {Task-Aware Machine Unlearning and Its Application in Load Forecasting},
  journal = {IEEE Transactions on Power Systems},
  volume  = {39},
  number  = {6},
  pages   = {7178--7189},
  year    = {2024},
  doi     = {10.1109/TPWRS.2024.3376828}
}

@inproceedings{koren2025membership,
  author    = {Koren, Noam and Goldsteen, Abigail and Amit, Guy and Farkash, Ariel},
  title     = {Membership Inference Attacks Against Time-Series Models},
  booktitle = {Proceedings of the 16th Asian Conference on Machine Learning},
  series    = {Proceedings of Machine Learning Research},
  volume    = {260},
  pages     = {319--334},
  year      = {2025}
}

\end{document}